\documentclass[final]{cvpr}

\usepackage{setspace}
\usepackage{times}
\usepackage{epsfig}
\usepackage{graphicx}
\usepackage{amsmath}
\usepackage{amssymb}
\usepackage{pifont}
\usepackage{color}
\newcommand{\tabincell}[2]{\begin{tabular}{@{}#1@{}}#2\end{tabular}}
\usepackage[numbers]{natbib}
\usepackage{multirow}
\usepackage{comment}
\usepackage[pagebackref=true,breaklinks=true,letterpaper=true,colorlinks,bookmarks=false]{hyperref}

\def\cvprPaperID{6156} 
\def\confYear{CVPR 2021}

\begin{document}
	
	\title{Exploring Sparsity in Image Super-Resolution for Efficient Inference}
	
	\author{Longguang Wang$^{1}$, Xiaoyu Dong$^{2,3}$, Yingqian Wang$^{1}$, Xinyi Ying$^{1}$, Zaiping Lin$^{1}$, Wei An$^{1}$, Yulan Guo$^{1*}$\\
		$^{1}$National University of Defense Technology~~~~~
		$^{2}$The University of Tokyo~~~~~~
		$^{3}$RIKEN AIP\\
		{\tt\small \{wanglongguang15,yulan.guo\}@nudt.edu.cn}
	}
	
	\maketitle
	\pagestyle{empty}
	\thispagestyle{empty}
	
	\begin{abstract}
		Current CNN-based super-resolution (SR) methods process all locations equally with computational resources being uniformly assigned in space. However, since missing details in low-resolution (LR) images mainly exist in regions of edges and textures, less computational resources are required for those flat regions.
		Therefore, existing CNN-based methods involve redundant computation in flat regions, which increases their computational cost and limits their applications on mobile devices. 
		\textcolor{black}{In this paper, we explore the sparsity in image SR to improve inference efficiency of SR networks.
		Specifically, we develop a Sparse Mask SR (SMSR) network to learn sparse masks to prune redundant computation.} Within our SMSR, spatial masks learn to identify ``important'' regions while channel masks learn to mark redundant channels in those ``unimportant'' regions. Consequently, redundant computation can be accurately localized and skipped while maintaining comparable performance. It is demonstrated that our SMSR achieves state-of-the-art performance with $41\%/33\%/27\%$ FLOPs being reduced for $\times2/3/4$ SR. \textcolor{black}{Code is available at: \url{https://github.com/LongguangWang/SMSR}.}
	\end{abstract}
	
	\section{Introduction}
	The goal of single image super-resolution (SR) is to recover a high-resolution (HR) image from a single low-resolution (LR) observation. Due to the powerful feature representation and model fitting capabilities of deep neural networks, CNN-based SR methods have achieved significant performance improvements over traditional ones. Recently, many efforts have been made towards real-world applications, including few-shot SR \cite{Shocher2018Zero,Soh2020Meta}, blind SR \cite{Gu2019Blind,Zhang2020Deep,Wang2021Unsupervised}, 
	and scale-arbitrary SR \cite{Hu2019Meta,Wang2020Learning}. \textcolor{black}{With the popularity of intelligent edge devices (such as smartphones and VR glasses), performing SR on these  devices is highly demanded.
	Due to the limited resources of edge devices\footnote{For example, the computational performance of Kirin 990 and RTX 2080Ti are 0.9 and 13.4 tFLOPS, respectively.}, efficient SR is crucial to the applications on these devices.}
	
	
	\begin{figure}[t]
		\centering 
		\includegraphics[width=0.77\linewidth]{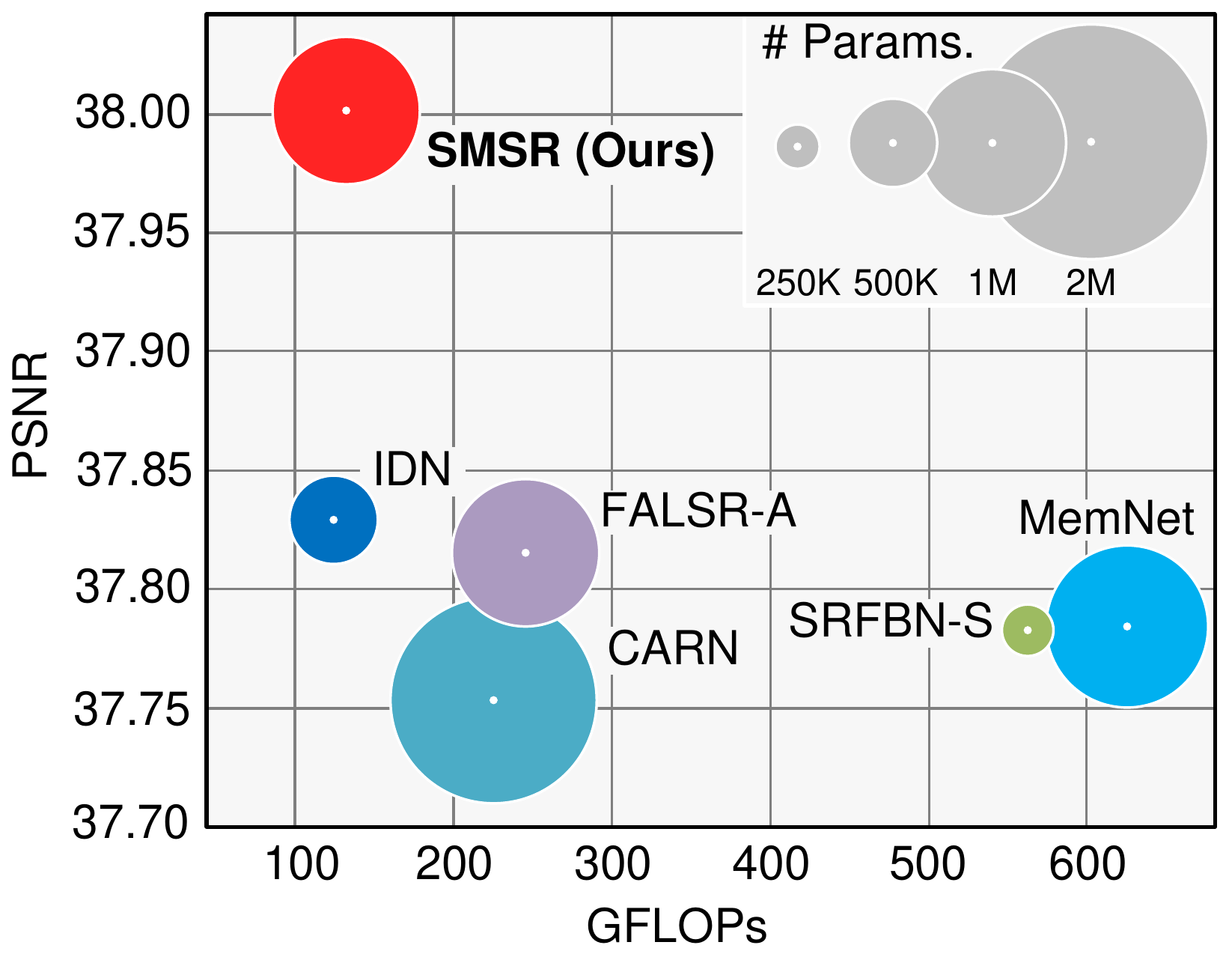}
		\vspace{-0.1cm}
		\caption{Trade-off between PSNR performance, number of parameters and FLOPs. Results are achieved on Set5 for $\times2$ SR. 
		}
		\label{fig5}
		\vspace{-0.6cm}
	\end{figure}
	
	Since the pioneering work of SRCNN \cite{Dong2014Learning}, deeper networks have been extensively studied for image SR. In VDSR \cite{Kim2016Accurate}, SR network is first deepened to 20 layers. Then, a very deep and wide architecture with over 60 layers is introduced in EDSR \cite{Lim2017Enhanced}. Later, Zhang \emph{et al.} further increased the network depth to over 100 and 400 in RDN \cite{Zhang2018Residual} and RCAN \cite{Zhang2018Image}, respectively. Although a deep network usually improves SR performance, it also leads to high computational cost and limits the applications on mobile devices. To address this problem, several efforts have been made to reduce model size through information distillation \cite{Hui2018Fast} and efficient feature reuse \cite{Ahn2018Fast}. 
	Nevertheless, these networks still involve redundant computation. 
	Compared to an HR image, missing details in its LR image mainly exist in regions of edges and textures. Consequently, less computational resources are required in those flat regions. However, these CNN-based SR methods process all locations equally, resulting in redundant computation within flat regions. 
	
	
	\textcolor{black}{In this paper, we explore the  sparsity in image SR to improve inference efficiency of SR networks.} We first study the intrinsic sparsity of the image SR task and then investigate the feature sparsity in existing SR networks. To fully exploit the sparsity for efficient inference, we propose a sparse mask SR (SMSR) network to dynamically skip redundant computation at a fine-grained level.  Our SMSR learns spatial masks to identify ``important'' regions (\emph{e.g.}, edge and texture regions) and uses channel masks to mark redundant channels in those ``unimportant'' regions. These two kinds of masks work jointly to accurately localize redundant computation. During network training, we soften these binary masks using the Gumbel softmax trick to make them differentiable. During inference, we use sparse convolution to skip redundant computation. It is demonstrated that our SMSR can effectively localize and prune redundant computation to achieve better efficiency while producing promising results (Fig.~\ref{fig5}).

	Our main contributions can be summarized as: 1) We develop an SMSR network to dynamically skip redundant computation for efficient image SR. In contrast to existing works that focus on lightweight network designs, we explore a different route by pruning redundant computation to improve inference efficiency.  
	2) We propose to localize redundant computation by learning spatial and channel masks. These two kinds of masks work jointly for fine-grained localization of redundant computation.
	3) Experimental results show that our SMSR achieves state-of-the-art performance with better inference efficiency. For example, our SMSR outperforms previous methods on Set14 for $\times2$ SR  with a significant speedup on mobile devices (Table~\ref{tab3}). 
	
	\begin{figure*}
		\centering 
		\begin{minipage}[u]{0.37\textwidth} 
			\includegraphics[width=0.83\textwidth]{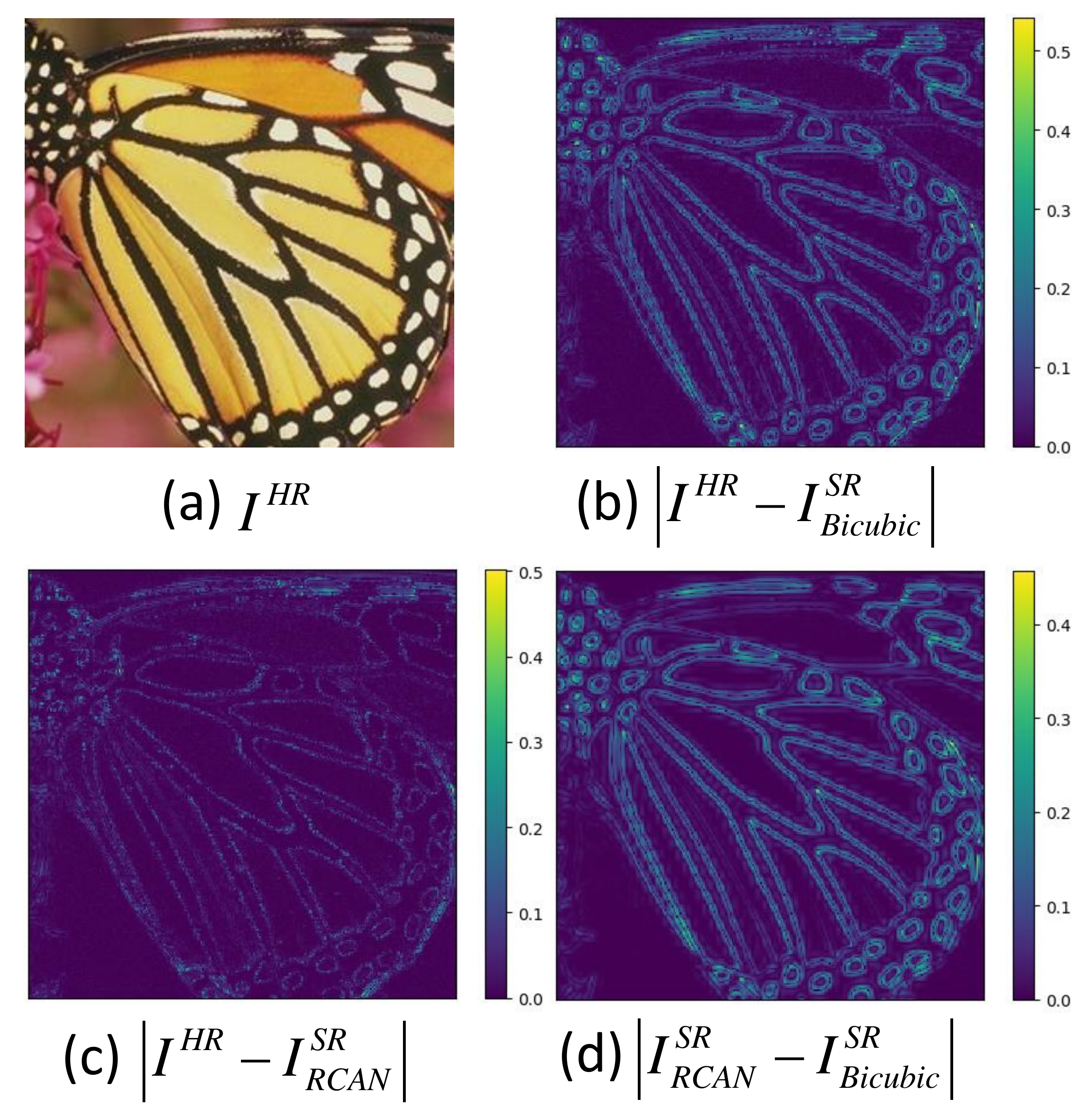}
			\caption{Absolute difference between $I^{SR}_{Bicubic}$, $I^{SR}_{RCAN}$ and $I^{HR}$ in the luminance channel.}
			\label{fig1}
		\end{minipage} 
		\hfill 
		\begin{minipage}[u]{0.57\textwidth} 
			\centering
			\includegraphics[width=0.9\textwidth]{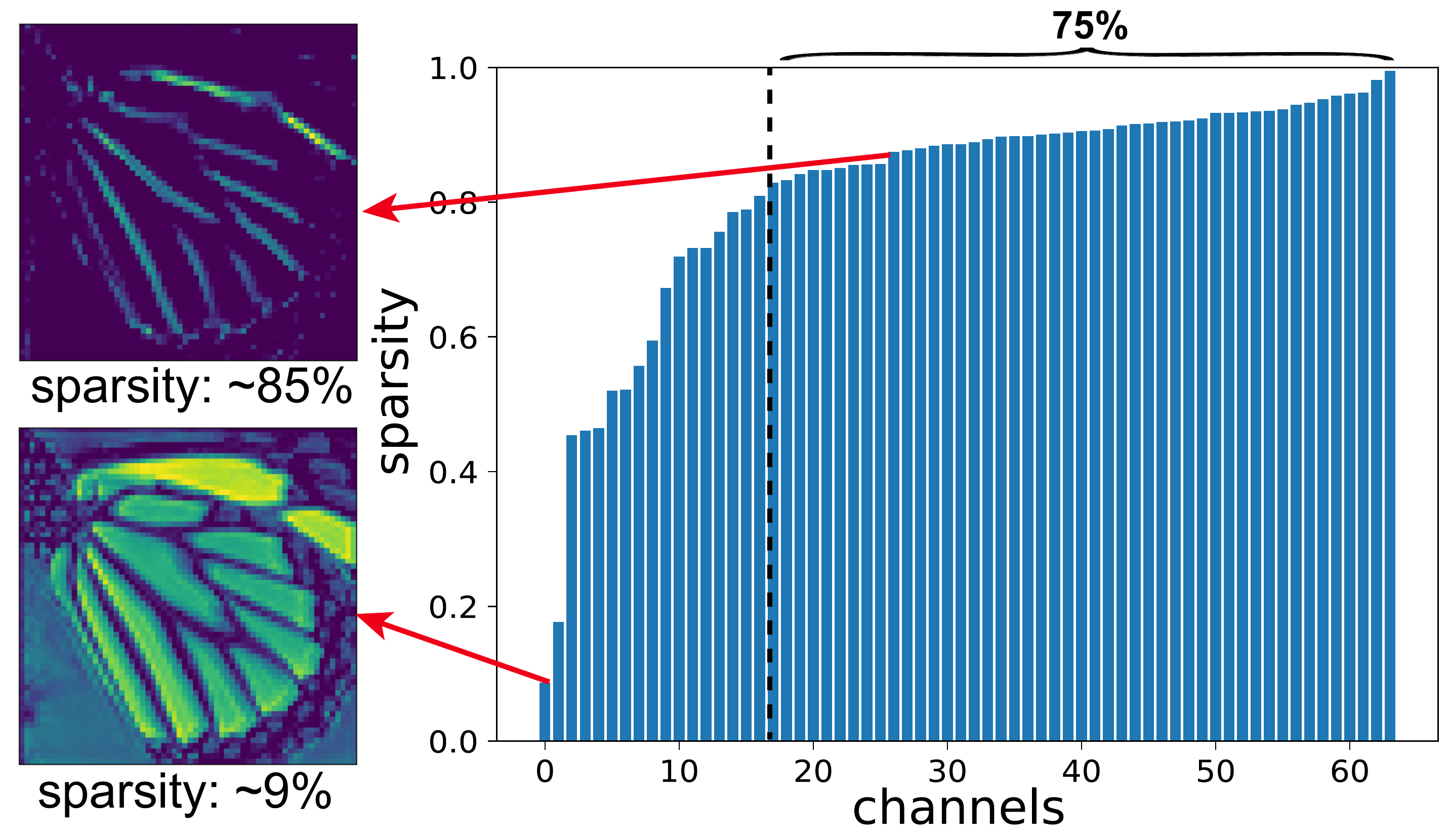} 
			\caption{Visualization of feature maps after the ReLU layer in the first backbone block of RCAN. Note that, sparsity is defined as the ratio of zeros in the corresponding channels.}
			\label{fig2}
		\end{minipage} 
		\vspace{-0.5cm}
	\end{figure*}
	
	\section{Related Work}
	In this section, we first review several major works for CNN-based single image SR. Then, we discuss CNN acceleration techniques related to our work, including adaptive inference and network pruning.
	
	\noindent{\textbf{Single Image SR.}}
	CNN-based methods have dominated the research of single image SR due to their strong representation and fitting capabilities. Dong \emph{et al.} \cite{Dong2014Learning} first introduced a three-layer network to learn an LR-to-HR mapping for single image SR. 
	Then, a deep network with 20 layers was proposed in VDSR \cite{Kim2016Accurate}.
	Recently, deeper networks are extensively studied for image SR. Lim \emph{et al.} \cite{Lim2017Enhanced} proposed a very deep and wide network (namely, EDSR) by cascading modified residual blocks. 
	Zhang \emph{et al.} \cite{Zhang2018Residual} further combined residual learning and dense connection to build RDN with over 100 layers. 
	Although these networks achieve state-of-the-art performance, the high computational cost and memory footprint limit their applications on mobile devices. 
	
	To address this problem, several lightweight networks were developed \cite{Lai2017Deep,Hui2018Fast,Ahn2018Fast}. Specifically, distillation blocks were proposed for feature learning in IDN \cite{Hui2018Fast}, while a cascading mechanism was introduced to encourage efficient feature reuse in CARN \cite{Ahn2018Fast}. Different from these manually designed networks, Chu \emph{et al.} \cite{Chu2020Fast} developed a compact architecture using neural architecture search (NAS). Recently, Lee \emph{et al.} \cite{Lee2020Learning} introduced a distillation framework to leverage knowledge learned by powerful teacher SR networks to boost the performance of lightweight student SR networks. Although these lightweight SR networks successfully reduce the model size, redundant computation is still involved and hinders them to achieve better computational efficiency.
	In contrast to many existing works that focus on compact architecture designs, few efforts have been made to exploit the redundancy in SR networks for efficient inference.
	
	\noindent{\textbf{Adaptive Inference.}}
	Adaptive inference techniques \cite{Wang2018SkipNet,Ren2018SBNet,Mullapudi2018HydraNets,Graham20183D,Li2019Improved} have attracted increasing interests since they can adapt the network structure according to the input. One active branch of adaptive inference techniques is to dynamically select an inference path at the levels of layers. Specifically, Wu \emph{et al.} \cite{Wu2018BlockDrop} proposed a BlockDrop approach for ResNets to dynamically drop several residual blocks for efficiency. Mullapudi \emph{et al.} \cite{Mullapudi2018HydraNets} proposed an HydraNet with multiple branches and used a gating approach to dynamically choose a set of them at test time. Another popular branch is early stopping techniques that skip the computation at a location whenever it is deemed to be unnecessary \cite{Xie2020Spatially}. On top of ResNets, Figurnov \emph{et al.} \cite{Figurnov2017Spatially} proposed a spatially adaptive computation time (SACT) mechanism to stop computation for a spatial position when the features become ``good enough''. 
	Liu \emph{et al.} \cite{Liu2020Deep} introduced adaptive inference for SR by producing a map of local network depth to adapt the number of convolutional layers implemented at different locations. However, these adaptive inference methods only focus on spatial redundancy without considering redundancy in channel dimension.
	
	\noindent{\textbf{Network Pruning.}}
	Network pruning \cite{Han2015Learning,Liu2017Learning,Luo2017ThiNet} is widely used to remove a set of redundant parameters for network acceleration. 
	As a popular branch of network pruning methods, structured pruning approaches are usually used to prune the network at the level of channels and even layers \cite{Li2017Pruning,Liu2017Learning,Luo2017ThiNet,He2019Filter}. 
	Specifically, Li \emph{et al.} \cite{Li2017Pruning} used $L_1$ norm to measure the importance of different filters and then pruned less important ones. Liu \emph{et al.} \cite{Liu2017Learning} imposed a sparsity constraint on  scaling factors of the batch normalization layers and identified channels with lower scaling factors as less informative ones.
	Different from these static structured pruning methods, Lin \emph{et al.} \cite{Lin2017Runtime} conducted runtime neural network pruning according to the input image. Recently, Gao \emph{et al.} \cite{Gao2019Dynamic} introduced a feature boosting and suppression method to dynamically prune unimportant channels at inference time. Nevertheless, these network pruning methods treat all spatial locations equally without taking their different importance into consideration. 

	\section{Sparsity in Image Super-Resolution}
	\label{sec3}
	In this section, we first illustrate the intrinsic sparsity of the single image SR task and then investigate the feature sparsity in state-of-the-art SR networks.
	
	Given an HR image $I^{HR}$ and its LR version $I^{LR}$ (\emph{e.g.}, $\times4$ downsampled), we super-resolve $I^{LR}$ using Bicubic and RCAN to obtain $I^{SR}_{Bicubic}$ and $I^{SR}_{RCAN}$, respectively. Figure~\ref{fig1} shows the absolute difference between $I^{SR}_{Bicubic}$, $I^{SR}_{RCAN}$ and $I^{HR}$ in the luminance channel. It can be observed from Fig.~\ref{fig1}(b) that $I^{SR}_{Bicubic}$ is ``good enough'' for flat regions, with noticeable missing details in only a small proportion of regions ($\sim\!17\%$ pixels with $|I^{HR}\!-\!I^{SR}_{Bicubic}|>0.1$). That is, the SR task is intrinsically sparse in spatial domain. Compared to Bicubic, RCAN performs better in edge regions while achieving comparable performance in flat regions (Fig.~\ref{fig1}(c)). Although RCAN focuses on recovering high-frequency details in edge regions (Fig.~\ref{fig1}(d)), those flat regions are equally processed at the same time. Consequently, redundant computation is involved.
	
	Figure~\ref{fig2} illustrates the feature maps after the ReLU layer in a backbone block of RCAN. It can be observed that the spatial sparsity varies significantly for different channels. Moreover, a considerable number of channels are quite sparse (sparsity $\geq$ 0.8), with only edge and texture regions being activated. That is, computation in those flat regions is redundant since these regions are not activated after the ReLU layer. In summary, RCAN activates only a few channels for ``unimportant'' regions (\emph{e.g.}, flat regions) and more channels for ``important'' regions (\emph{e.g.}, edge  regions). More results achieved with different SR networks and backbone blocks are provided in the supplemental material.
	
	Motivated by these observations, we learn sparse masks to localize and skip redundant computation for efficient inference. Specifically, our spatial masks dynamically identify ``important'' regions while the channel masks mark redundant channels in those ``unimportant'' regions. Compared to network pruning methods \cite{Gao2019Dynamic,Lin2017Runtime,He2019Filter}, we take region redundancy into consideration and only prune channels for ``unimportant'' regions. Different from adaptive inference networks \cite{Ren2018SBNet,Li2017Not}, we further investigate the redundancy in channel dimension to localize redundant computation at a finer-grained level.
	
	\begin{figure*}[t]
		\centering
		\includegraphics[width=0.85\linewidth]{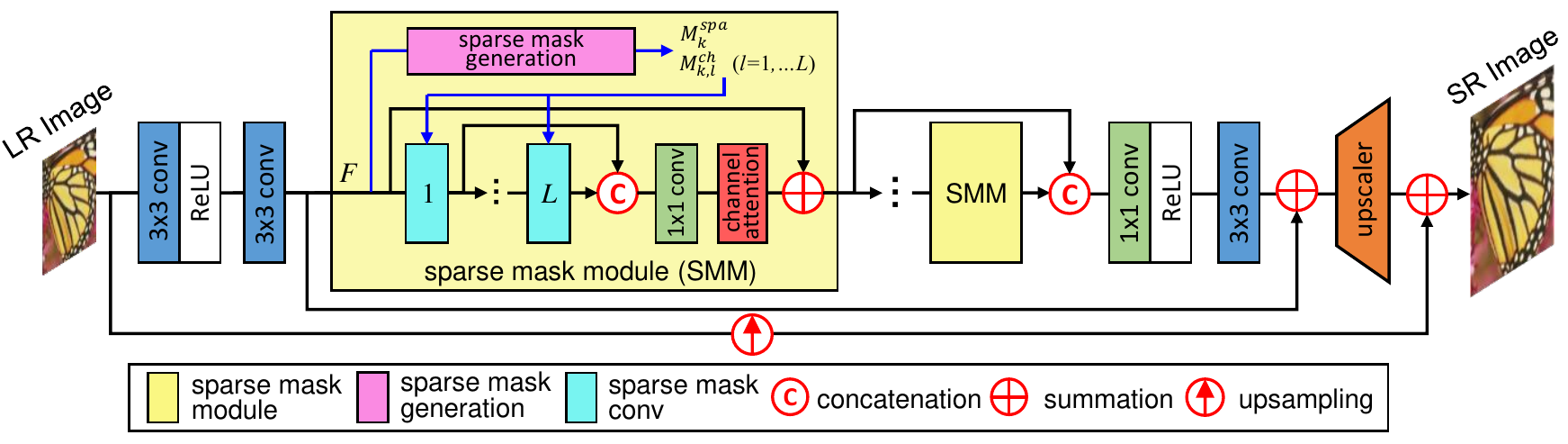}
		\caption{An overview of our SMSR network.}
		\vspace{-0.2cm}
		\label{fig3}
	\end{figure*}
	
	\section{Our SMSR Network}
	
	Our SMSR network uses sparse mask modules (SMM) to prune redundant computation for efficient image SR. Within each SMM, spatial and channel masks are first generated to localize redundant computation, as shown in Fig.~\ref{fig3}. \textcolor{black}{Then, the redundant computation is dynamically skipped using $L$ densely-connected sparse mask convolutions.} Since only necessary computation is performed, our SMSR can achieve better efficiency while maintaining comparable performance.
	
	\subsection{Sparse Mask Generation}
	
	\begin{figure*}[t]
		\centering
		\includegraphics[width=0.85\linewidth]{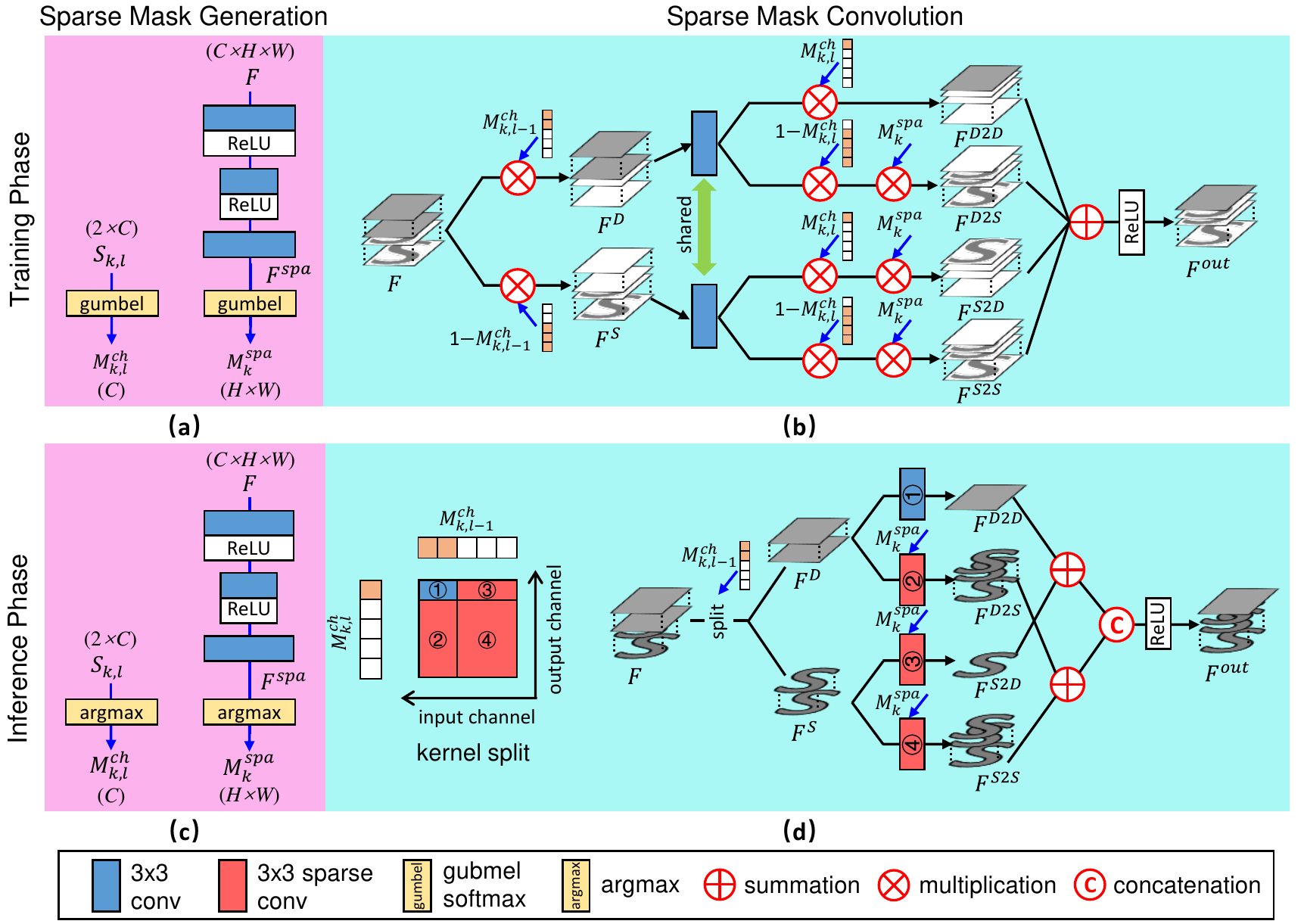}
		\caption{An illustration of sparse mask generation and sparse mask convolution.}
		\label{fig4}
		\vspace{-0.4cm}
	\end{figure*}

	\begin{spacing}{1.2}
		\noindent \textbf{1) Training Phase}
	\end{spacing}
	
	\noindent{\textbf{Spatial Mask.}}
	The goal of spatial mask is to identify ``important'' regions in feature maps (\emph{i.e.}, 0 for ``unimportant'' regions and 1 for ``important'' ones). 
	To make the binary spatial mask learnable, we use Gumbel softmax distribution to approximate the one-hot distribution \cite{Jang2017Categorical}.
	Specifically, input feature $F\!\in\!\mathbb{R}^{C\times{H}\times{W}}$ is first fed to an hourglass block to produce $F^{spa}\!\in\!\mathbb{R}^{2\times{H}\times{W}}$, as shown in Fig.~\ref{fig4}(a). Then, the Gumbel softmax trick  is used to obtain a softened spatial mask $M_k^{spa}\!\in\!\mathbb{R}^{{H}\times{W}}$:
	\vspace{-0.15cm}
	\begin{equation}
	M_k^{spa}[x,y]=\frac{{\rm exp}\Big(\big(F^{spa}[1,x,y]\!+\!G_k^{spa}[1,x,y]\big)/\tau\Big)}{\sum_{i=1}^2{\rm exp}\Big(\big(F^{spa}[i,x,y]\!+\!G_k^{spa}[i,x,y]\big)/\tau\Big)},
	\end{equation}
	where $x,y$ are vertical and horizontal indices, $G_k^{spa}\!\in\!\mathbb{R}^{2\!\times\!{H}\times\!{W}}$ is a Gumbel noise tensor with all elements following ${\rm Gumbel}(0,1)$ distribution and $\tau$ is a temperature parameter. 
	When $\tau\!\rightarrow\!\infty$, samples from Gumbel softmax distribution become uniform. That is, all elements in $M_k^{spa}$ are 0.5. When $\tau\!\rightarrow\!0$, samples from Gumbel softmax distribution become one-hot. That is, $M_k^{spa}$ becomes binary. In practice, we start at a high temperature and anneal to a small one to obtain binary spatial masks.
	
	\noindent{\textbf{Channel Mask.}}
	In addition to spatial masks, channel masks are used to mark redundant channels in those ``unimportant'' regions (\emph{i.e.}, 0 for redundant channels and 1 for preserved ones). Here, we also use Gumbel softmax trick to produce binary channel masks. For the $l^{\rm th}$ convolutional layer in the $k^{\rm th}$ SMM, we feed auxiliary parameter
	$S_{k,l}\in\mathbb{R}^{2\times{C}}$ to a Gumbel softmax layer to generate softened channel masks $M_{k,l}^{ch}\in\mathbb{R}^{{C}}$: 
	\vspace{-0.15cm}
	\begin{equation}
	\begin{aligned}
	M_{k,l}^{ch}[c]=\frac{{\rm exp}\Big(\big(S_{k,l}[1,c]+G_{k,l}^{ch}[1,c]\big)/\tau\Big)}{\sum_{i=1}^2{\rm exp}\Big(\big(S_{k,l}[i,c]+G_{k,l}^{ch}[i,c]\big)/\tau\Big)}
	\end{aligned},
	\end{equation}
	where $c$ is the channel index and $G_{k,l}^{ch}\in\mathbb{R}^{2\times{C}}$ is a Gumbel noise tensor. \textcolor{black}{In our experiments, $S_{k,l}$ is initialized using random values drawn from a Gaussian distribution ${\rm N}(0,1)$.}
	
	\begin{table*}
		\caption{Comparative results achieved on Set14 by our SMSR with different settings for $\times2$ SR.}
		\begin{center}
			\footnotesize
			\renewcommand\arraystretch{0.92}
			\setlength{\tabcolsep}{3mm}{
				\begin{tabular}{|c|c|c|c|c|c|c|c|c|}
					\hline 
					\multirow{1}{*}{Model} & \multirow{1}{*}{Spatial Mask} & \multirow{1}{*}{Channel Mask} & \multirow{1}{*}{Conv} & \multirow{1}{*}{\#Params.} 
					& Sparsity & \multirow{1}{*}{FLOPs} & \multirow{1}{*}{PSNR} & \multirow{1}{*}{SSIM} 
					\tabularnewline
					\hline
					1	& \ding{55}  & \ding{55} & Vanilla &926K & 0    &$1.00\times$&33.65&0.9180
					\tabularnewline
					2	& \ding{55}  & \ding{51} & Vanilla &587K & 0.46 &$0.60\times$&33.53&0.9169
					\tabularnewline
					3	& \ding{51}  & \ding{55} & Sparse  &985K & 0.42 &$0.65\times$&{33.60}&{0.9176}
					\tabularnewline
					4 (Ours)	& \ding{51}  & \ding{51} & Sparse &985K& 0.46 &$0.61\times$ &33.64&0.9179
					\tabularnewline
					\hline
			\end{tabular}}
		\end{center}
		\label{tab2}
		\vspace{-0.7cm}
	\end{table*}

	\noindent{\textbf{Sparsity Regularization.}}
	Based on spatial and channel masks, we define a sparsity term $\eta_{k,l}$:
	\vspace{-0.15cm}
	\begin{equation}
	\eta_{k,l}=\frac{1}{{C}\!\times\!H\!\times\!{W}}\sum_{c,x,y}
	\left(
	\begin{aligned}
	(1\!-\!M_{k,l}^{ch}[c])\!\times\!{M_k^{spa}[x,y]}\\
	+{M_{k,l}^{ch}[c]\!\times\!{I[x,y]}}
	\end{aligned}
	\right),
	\end{equation}
	where $I\!\in\!{\mathbb{R}^{{H}\times{W}}}$ is a tensor with all ones. Note that, $\eta_{k,l}$ represents the ratio of activated locations in the output feature maps. To encourage the output features to be more sparse with fewer locations being activated, we further introduce a sparsity regularization loss: 
	\vspace{-0.15cm}
	\begin{equation}
	\label{eq4}
	L_{reg}=\frac{1}{K\!\times\!{L}}\sum_{k,l}\eta_{k,l},
	\end{equation}
	where $K$ is the number of SMMs and $L$ is the number of sparse mask convolutional layers within each SMM.

	\noindent{\textbf{Training Strategy.}}
	During the training phase, the temperature parameter $\tau$ in Gumbel softmax layers is annealed using the schedule $\tau\!=\!{\rm max}(0.4,~~1\!-\!\frac{t}{T_{temp}})$, where $t$ is the number of epochs and $T_{temp}$ is empirically set to 500 in our experiments. As $\tau$ gradually decreases, Gumbel softmax distribution is forced to approach an one-hot distribution to produce binary spatial and channel masks.

	\begin{spacing}{1.2}
		\noindent \textbf{2) Inference Phase}
	\end{spacing}
	During training, Gumbel softmax distributions are forced to approach one-hot distributions as $\tau$ decreases. Therefore, we replace the Gumbel softmax layers with argmax layers after training to obtain binary spatial and channel masks, as shown in Fig.~\ref{fig4}(c). 

	\subsection{Sparse Mask Convolution}
	
	\begin{spacing}{1.2}
		\noindent \textbf{1) Training Phase}
	\end{spacing}
	
	To enable backpropagation of gradients at all locations, we do not explicitly perform sparse convolution during training. Instead, we multiply the results of a vanilla ``dense'' convolution with predicted spatial and channel masks, as shown in Fig.~\ref{fig4}(b). 
	Specifically, input feature $F$ is first multiplied with $M^{ch}_{k,l-1}$ and $(1\!-\!M^{ch}_{k,l-1})$ to obtain $F^D$ and $F^S$, respectively. That is, channels with ``dense'' and ``sparse'' feature maps in $F$ are \textcolor{black}{separated}. Next, $F^D$ and $F^S$ are passed to two convolutions with shared weights. The resulting features are then multiplied with different combinations of $(1\!-\!M^{ch}_{k,l})$, $M^{ch}_{k,l}$ and $M^{spa}_k$ to activate different parts of the features. Finally, all these features are summed up to generate the output feature $F^{out}$. 
	Thanks to Gumbel softmax trick used in mask generation, gradients at all locations can be preserved to optimize the kernel weights of convolutional layers.
	
	\begin{spacing}{1.2}
		\noindent \textbf{2) Inference Phase}
	\end{spacing}
	During the inference phase, sparse convolution is performed based on the predicted spatial and channel masks, as shown in Fig.~\ref{fig4}(d). Take the $l^{\rm th}$ layer in the $k^{\rm th}$ SMM as an example, its kernel is first splitted into four sub-kernels according to $M_{k,l-1}^{ch}$ and $M_{k,l}^{ch}$ to obtain four convolutions. 
	Meanwhile, input feature $F$ is splitted into $F^D$ and $F^S$ based on $M_{k,l-1}^{ch}$. Then, $F^D$ is fed to convolutions \ding{192} and \ding{193} to produce $F^{D2D}$ and $F^{D2S}$, while $F^S$ is fed to convolutions \ding{194} and \ding{195} to produce $F^{S2D}$ and $F^{S2S}$. Note that, $F^{D2D}$ is produced by a vanilla ``dense'' convolution while $F^{D2S}$, $F^{S2D}$ and $F^{S2S}$ are generated by sparse convolutions with only ``important'' regions (marked by $M_k^{spa}$) being computed. Finally, features obtained from these four branches are summed and concatenated to produce the output feature $F^{out}$. Using sparse mask convolution, computation for redundant channels within those ``unimportant'' regions can be skipped for efficient inference.
	
	\subsection{Discussion}
	Different from many recent works that use lightweight network designs \cite{Hui2018Fast,Ahn2018Fast,Chu2020Fast} or knowledge distillation \cite{Lee2020Learning} for efficient SR, we speedup SR networks by pruning redundant computation. Previous adaptive inference and network pruning methods focus on redundant computation in spatial and channel dimensions independently. Directly applying these approaches cannot fully exploit the redundancy in SR networks and suffers notable performance drop, as demonstrated in Sec.~\ref{Sec5.2}. In contrast, our SMSR provides a unified framework to consider redundancy in both spatial and channel dimensions. It is demonstrated that our spatial and channel masks are well compatible to each other and facilitate our SMSR to obtain fine-grained localization of redundant computation.

		\begin{table*}[t]
		\caption{Comparative results achieved on Set14 by our SMSR with different sparsities for $\times2$ SR.}
		\begin{center}
			\renewcommand\arraystretch{0.92}
			\footnotesize
			\setlength{\tabcolsep}{2mm}{
				\begin{tabular}{|l|c|c|c|c|c|c|cccc|c|c|}
					\hline 
					\multirow{2}{*}{Model} 
					& \multirow{2}{*}{Conv} 
					& \multirow{2}{*}{$\lambda_0$} 
					& \multirow{2}{*}{Sparsity}
					& \multirow{2}{*}{\#Params.}
					& \multirow{2}{*}{FLOPs} 
					& \multirow{2}{*}{Memory} 
					& \multicolumn{4}{c|}{Time}
					& \multirow{2}{*}{PSNR}
					& \multirow{2}{*}{SSIM}
					\tabularnewline
					&&&&&&&GPU & CPU & Kirin 990 & Kirin 810 & &
					\tabularnewline
					\hline
					
					baseline& Vanilla  & 0 & 0 & 926K & $1.00\times$ & $1.00\times$ & $1.00\times$ & $1.00\times$ & $1.00\times$ & $1.00\times$ 
					& 33.65 & 0.9180
					\tabularnewline
					5		& Sparse & 0.1 & 0.46 & 985K & $0.61\times$ & $0.89\times$ & $1.22\times$ & $0.79\times$ & $0.64\times$ & $0.57\times$ 
					& 33.64 & 0.9179
					\tabularnewline
					6		& Sparse & 0.2 & 0.64 & 985K & $0.46\times$ & $0.87\times$ & $1.11\times$ & $0.73\times$ & $0.55\times$ & $0.50\times$
					& 33.61 &0.9174
					\tabularnewline
					7		& Sparse & 0.3 & 0.73 & 985K & $0.38\times$ &$0.85\times$ & $1.04\times$ & $0.68\times$ & $0.54\times$ & $0.45\times$ 
					& 33.52 & 0.9169 
					\tabularnewline
					\hline
					IDN	\cite{Hui2018Fast} 		  & - & - & - & 553K & $0.57\times$ & $0.91\times$ & $1.04\times$ & $0.73\times$ & $0.71\times$ & $0.60\times$ 
					& 33.30 & 0.9148
					\tabularnewline		
					CARN \cite{Ahn2018Fast}		  & - & - & - & 1592K & $0.99\times$ & $1.01\times$ & $1.00\times$ & $0.89\times$ & $0.96\times$ & $1.15\times$
					& 33.52 & 0.9166
					\tabularnewline
					FALSR-A \cite{Chu2020Fast} & - & - & - & 1021K & $1.04\times$   & $2.02\times$ & $1.11\times$ & $1.05\times$ & $1.02\times$ &$0.92\times$ 
					& 33.55 & 0.9168
					\tabularnewline					
					\hline
			\end{tabular}}
		\end{center}
		\label{tab3}
		\vspace{-0.7cm}
	\end{table*}
	
	\begin{figure*}
		\centering 
		\begin{minipage}[r]{0.3\textwidth} 
			\includegraphics[width=0.8\textwidth]{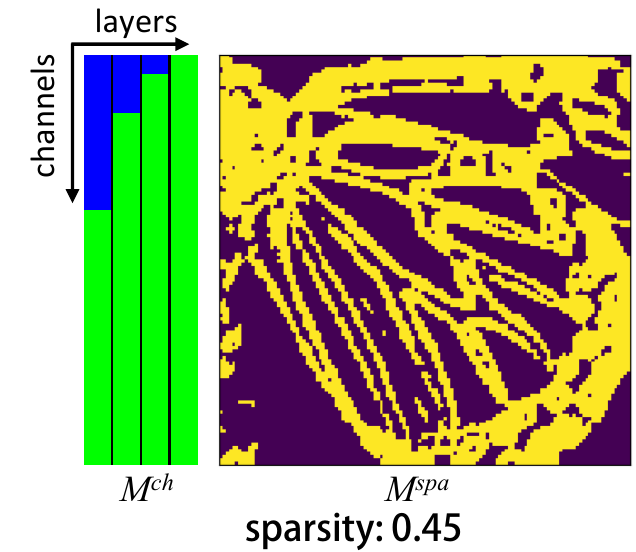}
			\caption{Visualization of sparse masks. Blue and green regions in $M^{ch}$ represent channels with ``dense'' and ``sparse'' feature maps, respectively. In $M^{spa}$, ``important'' locations are shown in yellow.}
			\label{fig7}
		\end{minipage} 
		\hfill 
		\begin{minipage}[c]{0.32\textwidth} 
			\centering
			\includegraphics[width=1.01\textwidth]{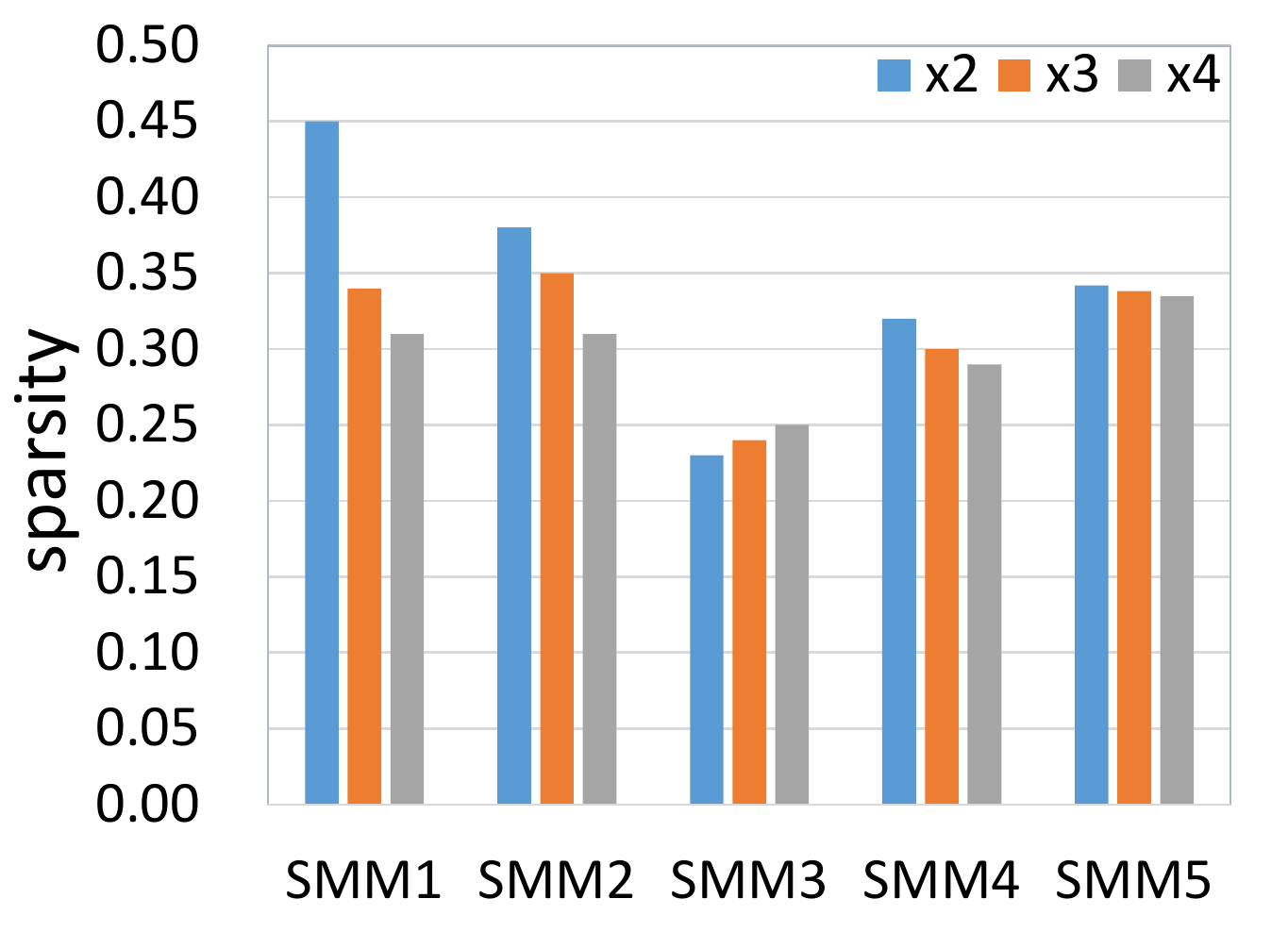} 
			\caption{Comparison of sparsities achieved in different SMMs on \emph{butterfly} for different scale factors.}
			\label{fig9}
		\end{minipage} 
		\hfill
		\begin{minipage}[c]{0.32\textwidth} 
			\centering
			\includegraphics[width=0.9\textwidth]{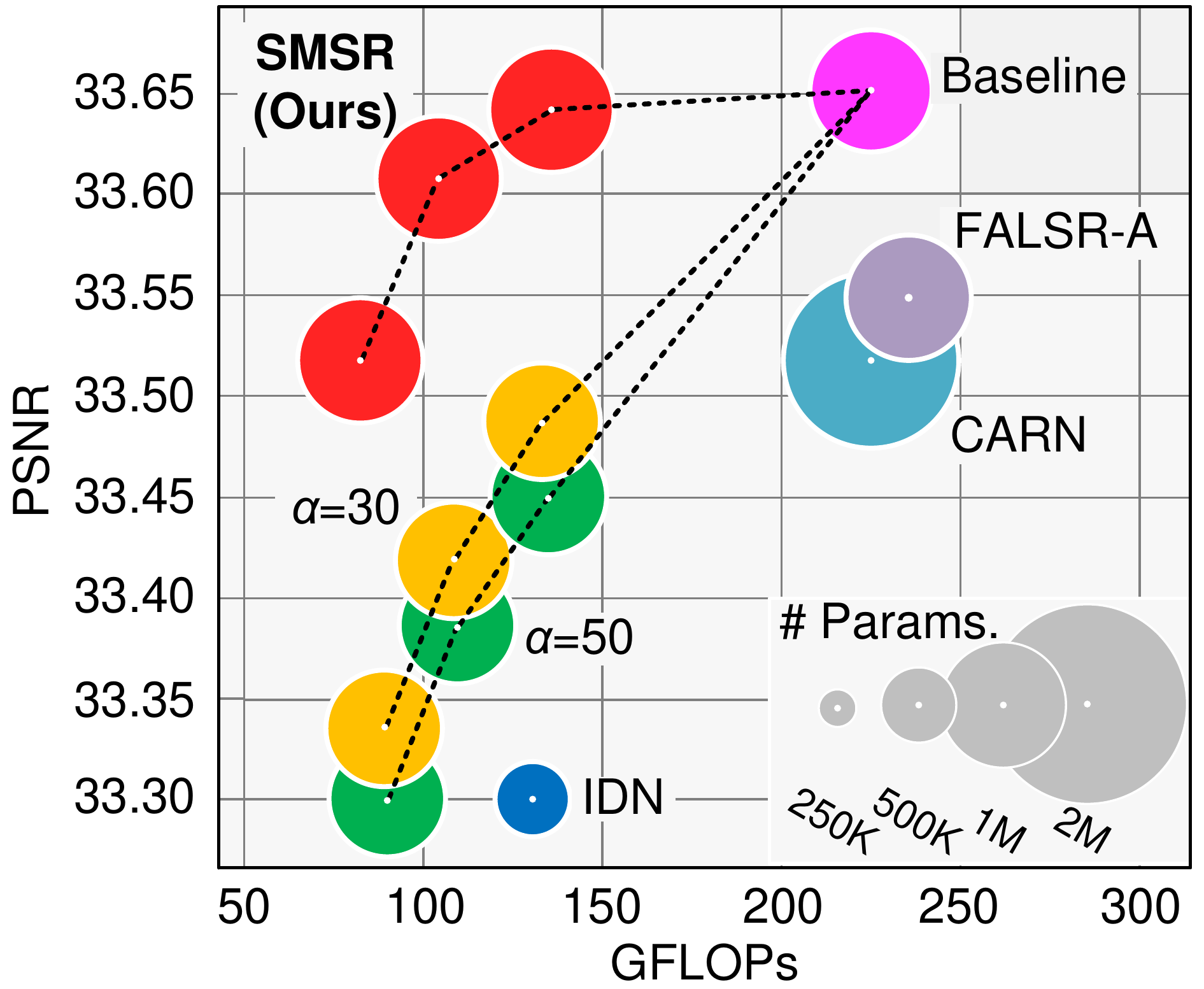} 
			\caption{Comparison between learning-based masks (red circles) and gradient-based masks (yellow and green circles) on Set14.}
			\label{fig10}
		\end{minipage} 
		\vspace{-0.4cm}
	\end{figure*}

	\section{Experiments}
	
	\subsection{Implementation Details}
	We used 800 training images and 100 validation images from the DIV2K dataset \cite{Agustsson2017NTIRE} as training and validation sets. For evaluation, we used five benchmark datasets  including Set5 \cite{Bevilacqua2012Low}, Set14 \cite{Zeyde2010Single}, B100 \cite{Martin2001database}, Urban100 \cite{Huang2015Single}, and Manga109 \cite{Matsui2017Sketch}. Peak signal-to-noise ratio (PSNR) and structural similarity index (SSIM) were used as evaluation metrics to measure SR performance. Following the evaluation protocol in \cite{Zhang2018Image,Zhang2018Residual}, we cropped borders and calculated the metrics in the luminance channel.

	During training, 16 LR patches of size $96\times96$ and their corresponding HR patches were randomly cropped. Data augmentation was then performed through random rotation and flipping. We set $C\!=\!64,L\!=\!4,K\!=\!5$ for our SMSR. We used the Adam method \cite{Kingma2015Adam} with $\beta_{1}=0.9$ and $\beta_{2}=0.999$ for optimization. The initial learning rate was set to $2\times10^{-4}$ and reduced to half after every 200 epochs. The training was stopped after 1000 epochs. The overall loss for training is defined as $L=L_{SR}+\lambda{L_{reg}}$, where $L_{SR}$ is the $L_1$ loss between SR results and HR images, $L_{reg}$ is defined in Eq.~\ref{eq4}. To maintain training stability, we used a warmup strategy $\lambda=\lambda_0\times{\rm min}(\frac{t}{T_{warm}}, 1)$,	where $t$ is the number of epochs, $T_{warm}$ is empirically set to 50 and $\lambda_0$ is set to 0.1.

	\subsection{Model Analysis}
	\label{Sec5.2}
	We first conduct experiments to demonstrate the effectiveness of sparse masks. Then, we investigate the effect of sparsity and visualize sparse masks for discussion. Finally, we compare our learning-based masks with heuristic ones.

	\noindent{\textbf{Effectiveness of Sparse Masks.}}
	To demonstrate the effectiveness of our sparse masks, we first introduced variant 1 by removing both spatial and channel masks. Then, we developed variants 2 and 3 by adding channel masks and spatial masks, respectively. Comparative results are shown in Table~\ref{tab2}. Without spatial and channel masks,  all locations and all channels are processed equally. Therefore, variant 1 has a high computational cost. Using channel masks,  redundant channels are pruned at all spatial locations. Therefore, variant 2  can be considered as a pruned version of variant 1. Although variant 2 has fewer parameters and FLOPs, it suffers a notable performance drop (33.53 vs. 33.65) since beneficial information in ``important'' regions of these pruned channels are discarded. With only spatial masks, variant 3 suffers from a conflict between efficiency and performance since 
	redundant computation in channel dimension
	cannot be well handled. Consequently, its FLOPs is reduced with a performance drop (33.60 vs. 33.65). 
	Using both spatial and channel masks, our SMSR can effectively localize and skip redundant computation at a finer-grained level to reduce FLOPs by $39\%$ while maintaining comparable performance (33.64 vs. 33.65).
	
	\noindent{\textbf{Effect of Sparsity.}}
	To investigate the effect of sparsity, we retrained our SMSR with large $\lambda_0$ to encourage high sparsity. Nvidia RTX 2080Ti, Intel I9-9900K and Kirin 990/810 were used as platforms of GPU, CPU and mobile processor for evaluation. For fair comparison of memory consumption and inference time, all convolutional layers in the backbone of different networks were implemented using im2col \cite{chellapilla2006high} based convolutions since different implementation methods (\emph{e.g.}, Winograd \cite{Lavin2016Fast} and FFT \cite{Vasilache2015Fast}) have different computational costs. Comparative results are presented in Table~\ref{tab3}.

	As $\lambda_0$ increases, our SMSR produces higher sparsities with more FLOPs and memory consumption being reduced. Further, our network also achieves significant speedup on CPU and mobile processors. 
	Due to the irregular and fragmented memory patterns, sparse convolution cannot make full use of the characteristics of general GPUs (\emph{e.g.}, memory coalescing) and relies on specialized designs to improve memory locality and cache hit rate for acceleration \cite{Yu2017Scalpel}.
	Therefore, the advantage of our SMSR cannot be fully exploited on GPUs without specific optimization. 
	Compared to other state-of-the-art methods, our SMSR (variant 5) obtains better performance with lower memory consumption and shorter inference time on  mobile processors. This clearly demonstrates the great potential of our SMSR for applications on mobile devices.

	
	
	\begin{table*}[t]
		\setcounter{table}{3}
		\caption{Comparative results achieved for $\times2/3/4$ SR. PSNR/SSIM results of previous works are directly copied from corresponding papers. FLOPs is computed based on HR images with a resolution of 720p ($1280\times720$). For SMSR, average sparsities on all datasets ($0.49/0.39/0.33$ for $\times2/3/4$ SR) are used to calculate FLOPs, with full FLOPs being shown in brackets. Best and second best results are \textbf{highlighted} and \underline{underlined}.}
		\footnotesize
		\vspace{-0.3cm}
		\begin{center}
			\renewcommand\arraystretch{0.92}
			\setlength{\tabcolsep}{2.6mm}{
				\begin{tabular}{|l|c|r|r|c|c|c|c|c|c|c|}
					\hline 
					Model & Scale & \#Params & FLOPs & Set5 & Set14 & B100 & Urban100 & Manga109
					\tabularnewline
					\hline
					\hline
					Bicubic 
					& $\times2$ & - & - & 33.66/0.9299 & 30.24/0.8688 & 29.56/0.8431 & 26.88/0.8403 & 30.80/0.9339
					\tabularnewline
					SRCNN \cite{Dong2014Learning} 	
					& $\times2$ & 57K & 52.7G & 36.66/0.9542 & 32.45/0.9067 & 31.36/0.8879 & 29.50/0.8946 & 35.60/0.9663
					\tabularnewline
					VDSR \cite{Kim2016Accurate} 	
					& $\times2$ & 665K & 612.6G & 37.53/0.9590 & 33.05/0.9130 & 31.90/0.8960 & 30.77/0.9140 & 37.22/0.9750
					\tabularnewline
					DRCN \cite{Kim2016Deeply} 	
					& $\times2$ & 1774K & 9788.7G & 37.63/0.9588 & 33.04/0.9118 & 31.85/0.8942 & 30.75/0.9133 & 37.55/0.9732
					\tabularnewline
					LapSRN \cite{Lai2017Deep} 		
					& $\times2$ & 813K & 29.9G & 37.52/0.9591 & 33.08/0.9130 & 31.08/0.8950 & 30.41/0.9101 & 37.27/0.9740
					\tabularnewline
					MemNet \cite{Tai2017MemNet} 	
					& $\times2$ & 677K & 623.9G & 37.78/0.9597 & 33.28/0.9142 & 32.08/0.8978 & 31.31/0.9195 & 37.72/0.9740
					\tabularnewline
					SRFBN-S \cite{Li2018Feedback} 		
					& $\times2$ & 282K & 574.4G & 37.78/0.9597 & 33.35/0.9156 & 32.00/0.8970 & 31.41/0.9207 & 38.06/0.9757
					\tabularnewline
					IDN \cite{Hui2018Fast} 		
					& $\times2$ & 553K & 127.7G & \underline{37.83/0.9600} & 33.30/0.9148 & 32.08/0.8985 & 31.27/0.9196 & 38.01/0.9749
					\tabularnewline
					CARN \cite{Ahn2018Fast} 		
					& $\times2$ & 1592K & 222.8G & 37.76/0.9590 & 33.52/0.9166 & 32.09/0.8978 & 31.92/0.9256 & \underline{38.36/0.9765}
					\tabularnewline
					FALSR-A \cite{Chu2020Fast} 		
					& $\times2$ & 1021K & 234.7G & 37.82/0.9595 & \underline{33.55/0.9168} & \underline{32.12/0.8987} & \underline{31.93/0.9256} & -/-
					\tabularnewline
					SMSR 		
					& $\times2$ &985K & \tabincell{r}{(224.1G)131.6G} & \textbf{38.00/0.9601} & \textbf{33.64/0.9179} & \textbf{32.17/0.8990} & \textbf{32.19/0.9284} & \textbf{38.76/0.9771} 
					\tabularnewline
					\hline
					\hline
					Bicubic 						
					& $\times3$ & - & - & 30.39/0.8682 & 27.55/0.7742 & 27.21/0.7385 & 24.46/0.7349 & 26.95/0.8556 
					\tabularnewline
					SRCNN \cite{Dong2014Learning} 	
					& $\times3$ & 57K & 52.7G & 32.75/0.9090 & 29.30/0.8215 & 28.41/0.7863 & 26.24/0.7989 & 30.48/0.9117
					\tabularnewline
					VDSR \cite{Kim2016Accurate}	 	
					& $\times3$ & 665K & 612.6G & 33.67/0.9210 & 29.78/0.8320 & 28.83/0.7990 & 27.14/0.8290 & 32.01/0.9340
					\tabularnewline
					DRCN \cite{Kim2016Deeply}	 	
					& $\times3$ & 1774K & 9788.7G & 33.82/0.9226 & 29.76/0.8311 & 28.80/0.7963 & 27.14/0.8279 & 32.24/0.9343
					\tabularnewline
					MemNet \cite{Tai2017MemNet} 	
					& $\times3$ & 677K & 623.9G & 34.09/0.9248 & 30.01/0.8350 & 28.96/0.8001 & 27.56/0.8376 & 32.51/0.9369
					\tabularnewline
					SRFBN-S \cite{Li2018Feedback} 		
					& $\times3$ & 375K & 686.4G & 34.20/0.9255 & 30.10/0.8372 & 28.96/0.8010 & 27.66/0.8415 & 33.02/0.9404
					\tabularnewline
					IDN \cite{Hui2018Fast} 		
					& $\times3$ & 553K & 57.0G & 34.11/0.9253 & 29.99/0.8354 & 28.95/0.8013 & 27.42/0.8359 & 32.71/0.9381
					\tabularnewline
					CARN \cite{Ahn2018Fast} 		
					& $\times3$ & 1592K & 118.8G & \underline{34.29/0.9255} & \underline{30.29/0.8407} & \underline{29.06/0.8034} & \underline{28.06/0.8493} & \underline{33.50/0.9440}
					\tabularnewline
					SMSR		
					& $\times3$ & 993K & \tabincell{r}{(100.5G)67.8G} & \textbf{34.40/0.9270} & \textbf{30.33/0.8412} & \textbf{29.10/0.8050} & \textbf{28.25/0.8536} & \textbf{33.68/0.9445} 
					\tabularnewline
					\hline
					\hline
					Bicubic 						
					& $\times4$ & - & -  & 28.42/0.8104 & 26.00/0.7027 & 25.96/0.6675 & 23.14/0.6577 & 24.89/0.7866
					\tabularnewline
					SRCNN \cite{Dong2014Learning} 	
					& $\times4$ & 57K & 52.7G & 30.48/0.8628 & 27.50/0.7513 & 26.90/0.7101 & 24.52/0.7221 & 27.58/0.8555
					\tabularnewline
					VDSR \cite{Kim2016Accurate} 	
					& $\times4$ & 665K & 612.6G  & 31.35/0.8830 & 28.02/0.7680 & 27.29/0.7260 & 25.18/0.7540 & 28.83/0.8870
					\tabularnewline
					DRCN \cite{Kim2016Deeply}	 	
					& $\times4$ & 1774K & 9788.7G & 31.53/0.8854 & 28.02/0.7670 & 27.23/0.7233 & 25.18/0.7524 & 28.93/0.8854
					\tabularnewline
					LapSRN \cite{Lai2017Deep} 		
					& $\times4$ & 813K & 149.4G & 31.54/0.8850 & 28.19/0.7720 & 27.32/0.7270 & 25.21/0.7560 & 29.09/0.8900
					\tabularnewline
					MemNet \cite{Tai2017MemNet} 	
					& $\times4$ & 677K & 623.9G & 31.74/0.8893 & 28.26/0.7723 & 27.40/0.7281 & 25.50/0.7630 & 29.42/0.8942
					\tabularnewline
					SRFBN-S \cite{Li2018Feedback} 		
					& $\times4$ & 483K &852.9G & 31.98/0.8923 & 28.45/0.7779 & 27.44/0.7313 & 25.71/0.7719 & 29.91/0.9008
					\tabularnewline
					IDN \cite{Hui2018Fast} 		
					& $\times4$ & 553K & 32.3G & 31.82/0.8903 & 28.25/0.7730 & 27.41/0.7297 & 25.41/0.7632 & 29.41/0.8942
					\tabularnewline
					CARN \cite{Ahn2018Fast} 		
					& $\times4$ & 1592K & 90.9G & \textbf{32.13/0.8937} & \textbf{28.60}/\underline{0.7806} & \textbf{27.58}/\underline{0.7349} & \underline{26.07/0.7837} & \underline{30.47/0.9084}
					\tabularnewline
					SMSR 	
					& $\times4$ &1006K&\tabincell{r}{(57.2G)41.6G} & \underline{32.12/0.8932} & \underline{28.55}/\textbf{0.7808} & \underline{27.55}/\textbf{0.7351} & \textbf{26.11/0.7868} & \textbf{30.54/0.9085}
					\tabularnewline
					\hline
			\end{tabular}}
		\end{center}
		\label{tab1}
		\vspace{-0.6cm}
	\end{table*}

	\begin{table}
		\setcounter{table}{2}
		\caption{Comparison between learning-based masks and gradient-based masks. Results are achieved on Set14 for $\times2$ SR.}
		\vspace{-0.3cm}
		\begin{center}
			\renewcommand\arraystretch{0.92}
			\footnotesize
			\setlength{\tabcolsep}{2.2mm}{
				\begin{tabular}{|c|c|c|c|cc|}
					\hline 
					\multirow{2}{*}{$M^{spa}$} & \multirow{2}{*}{\#Params.} & \multirow{2}{*}{$\alpha$} & \multirow{2}{*}{Sparsity} & \multicolumn{2}{c|}{Set14} 
					\tabularnewline
					& & & & PSNR & SSIM 
					\tabularnewline
					\hline
					\multirow{6}{*}{Gradient-based} 
					& 926K & 30 & 0.51 & 33.48 & 0.9163
					\tabularnewline
					& 926K & 30 & 0.62 & 33.42 & 0.9155
					\tabularnewline 
					& 926K & 30 & 0.72 & 33.33 & 0.9151
					\tabularnewline
					\cline{2-6}
					& 926K & 50 & 0.50 & 33.45 & 0.9162
					\tabularnewline
					& 926K & 50 & 0.61 & 33.39 & 0.9153
					\tabularnewline
					& 926K & 50 & 0.71 & 33.30 & 0.9150
					\tabularnewline
					\hline
					\multirow{3}{*}{Learning-based} 
					& 985K & - & 0.46 & 33.64 & 0.9179
					\tabularnewline
					& 985K & - & 0.64 & 33.61 & 0.9174
					\tabularnewline
					& 985K & - & 0.73 & 33.52 & 0.9169
					\tabularnewline
					\hline
			\end{tabular}}
		\end{center}
		\label{tab4}
		\vspace{-0.8cm}
	\end{table}

	\noindent{\textbf{Visualization of Sparse Masks.}}
	We visualize the sparse masks generated in the first SMM for $\times2$ SR in Fig.~\ref{fig7}. More results are provided in the supplemental material. It can be seen that locations around edges and textures in $M^{spa}$ are considered as ``important'' ones, which is consistent with our observations in Sec.~\ref{sec3}. Moreover, we can see that there are more sparse channels (\emph{i.e.}, green regions in $M^{ch}$)  in deep layers than shallow layers. This means that a subset of channels in shallow layers are informative enough for ``unimportant'' regions and our network progressively focuses more on ``important'' regions as the depth increases. Overall, our spatial and channel masks work jointly for fine-grained localization of redundant computation.
	
	We further investigate the sparsities achieved by our SMMs for different scale factors. Specifically, we feed an LR image ($\times$2 downsampled) to $\times2/3/4$ SMSR networks and compare the sparsities in their SMMs. As shown in Fig.~\ref{fig9}, the sparsities decrease for larger scale factors in most SMMs. Since more details need to be reconstructed for larger scale factors, more locations are marked as ``important'' ones (with sparsities being decreased).

	\noindent{\textbf{Learning-based Masks vs. Heuristic Masks.}} As regions of edges are usually identified as important ones in our spatial masks (Fig.~\ref{fig7}), another straightforward choice is to use heuristic masks. KernelGAN \cite{Bell-Kligler2019Blind} follows this idea to identify regions with large gradients as important ones when applying ZSSR \cite{Shocher2018Zero} and uses a masked loss to focus on these regions. To demonstrate the effectiveness of learning-based masks in our SMSR, we introduced a variant with gradient-induced masks. Specifically, we consider locations with gradients larger than a threshold $\alpha$ as important ones and keep the spatial mask fixed within the network. 
	The performance of this variant is compared to our SMSR in Table~\ref{tab4}. 
	Compared to learning-based masks, the variant with gradient-based masks suffers a notable performance drop with comparable sparsity (\emph{e.g.}, 33.52 vs. 33.33/33.30). 
	Further, we can see from Fig.~\ref{fig10}  that learning-based masks facilitate our SMSR to achieve better trade-off between SR performance and computational efficiency.
	Using fixed heuristic masks, it is difficult to obtain fine-grained localization of redundant computation. In contrast, learning-based masks enable our SMSR to accurately localize redundant computation to produce better results.
	
	\begin{figure*}[t]
		\centering
		\includegraphics[width=0.82\linewidth]{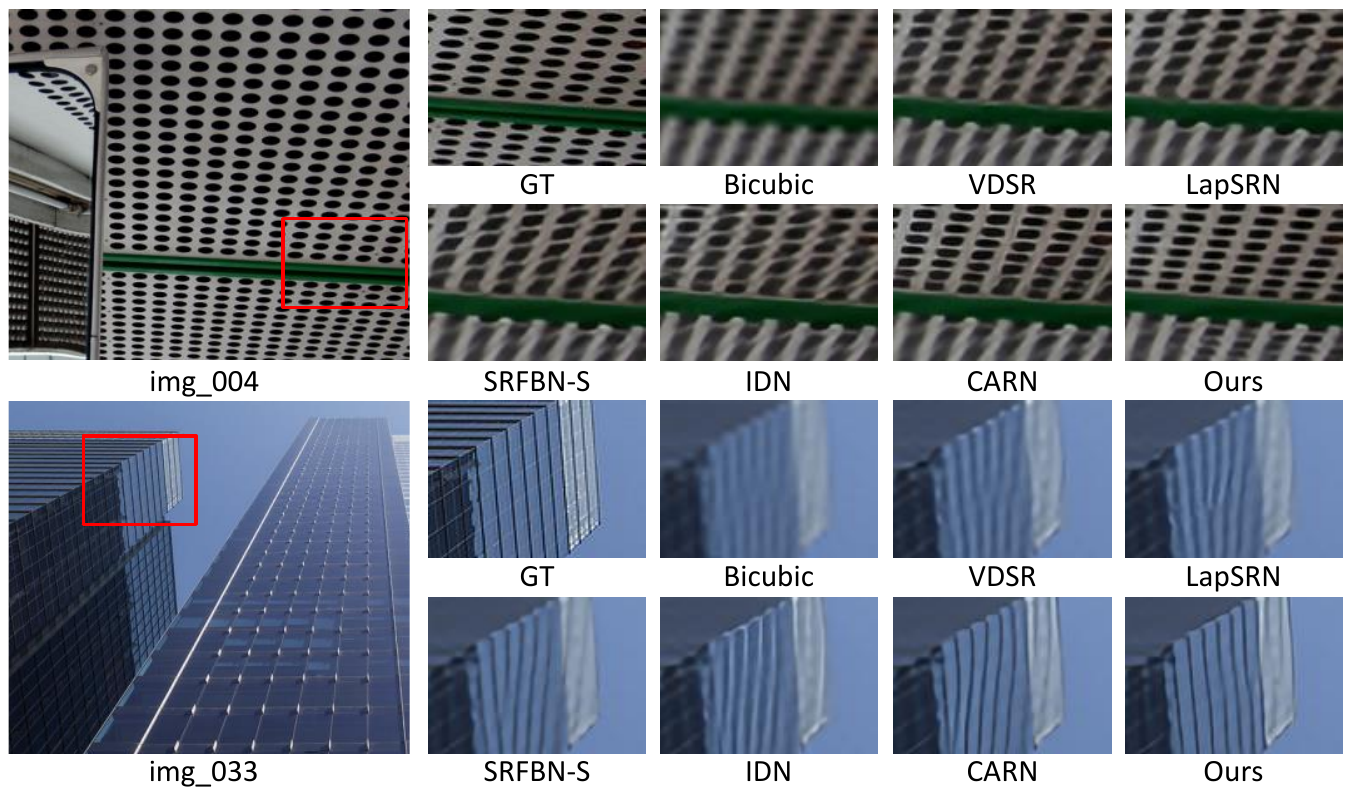}
		\vspace{-0.1cm}
		\caption{Visual comparison on the Urban100 dataset for $\times4$ SR.} 
		\label{fig6}
		\vspace{-0.3cm}
	\end{figure*}
	
	\begin{figure*}
		\centering
		\includegraphics[width=0.82\linewidth]{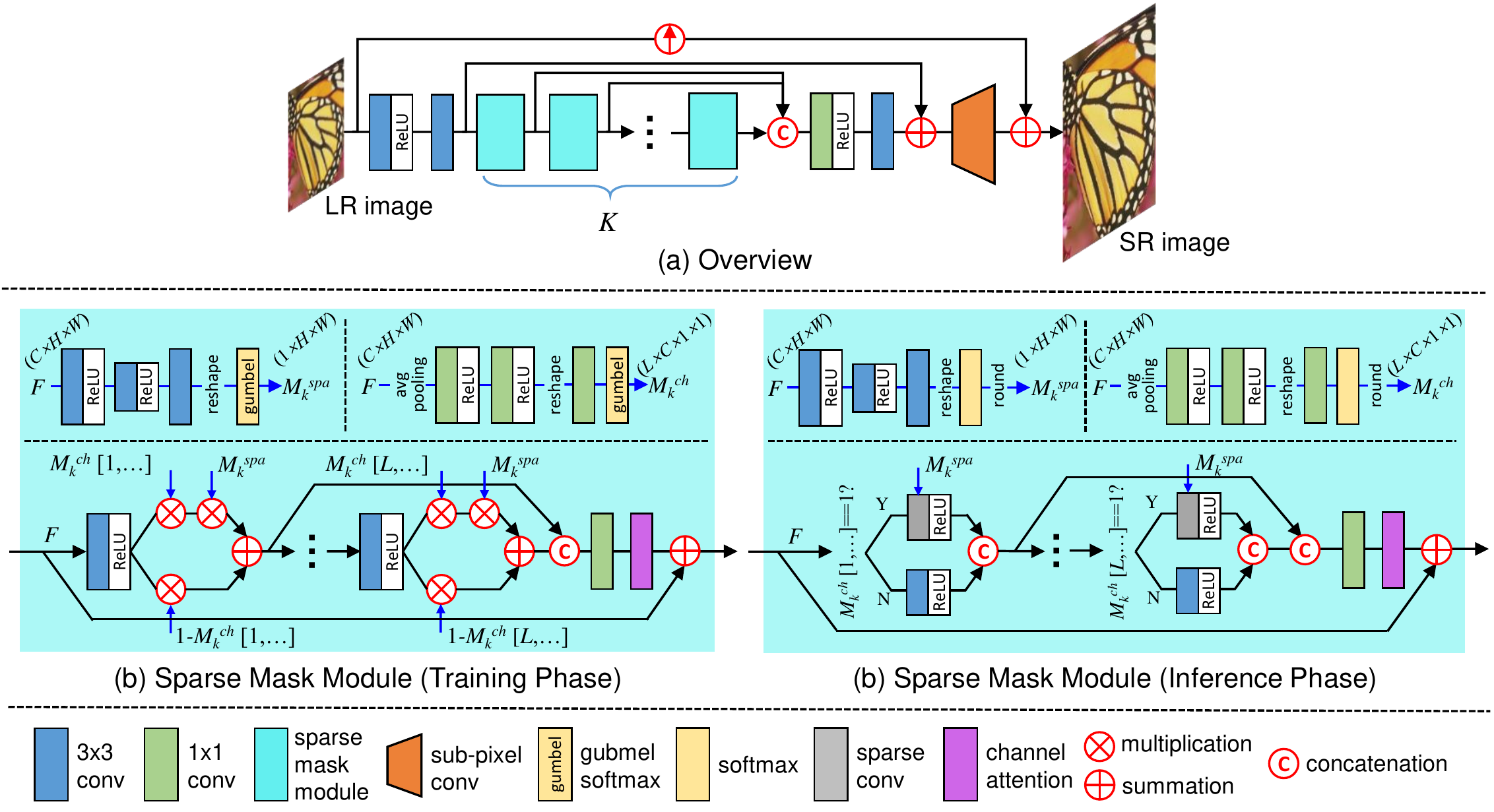}
		\vspace{-0.2cm}
		\caption{Visual comparison on a real-world image.} 
		\label{fig8}
		\vspace{-0.4cm}
	\end{figure*}
	
	\subsection{Comparison with State-of-the-art Methods}
	We compare our SMSR with nine state-of-the-art methods, including SRCNN \cite{Dong2014Learning}, VDSR \cite{Kim2016Accurate}, DRCN \cite{Kim2016Deeply}, LapSRN \cite{Lai2017Deep}, MemNet \cite{Tai2017MemNet}, SRFBN-S \cite{Li2018Feedback}, IDN \cite{Hui2018Fast}, CARN \cite{Ahn2018Fast}, and FALSR-A \cite{Chu2020Fast}. As this paper focuses on lightweight SR networks ($<2$M), several recent works with large models (\emph{e.g.}, EDSR \cite{Lim2017Enhanced} ($\sim$40M), RCAN \cite{Zhang2018Image} ($\sim$15M) and SAN \cite{Dai2019Second} ($\sim$15M)) are not included for comparison. Quantitative results are presented in Table~\ref{tab1} and visualization results are shown in Figs.~\ref{fig6} and \ref{fig8}.

	\noindent{\textbf{Quantitative Results.}}
	As shown in Table~\ref{tab1}, our SMSR outperforms the state-of-the-art methods on most datasets. For example, our SMSR achieves much better performance than CARN for $\times2$ SR, with the number of parameters and FLOPs being reduced by 38\% and 41\%, respectively. With a comparable model size, our SMSR performs favorably against FALSR-A and achieves better inference efficiency in terms of FLOPs (131.6G vs. 234.7G). 
	With comparable computational complexity in terms of FLOPs (131.6G vs. 127.7G), our SMSR achieves much higher PSNR values than IDN. Using sparse masks to skip redundant computation, our SMSR reduces $41\%/33\%/27\%$ FLOPs for $\times2/3/4$ SR while maintaining the state-of-the-art performance. We further show the trade-off between performance, number of parameters and FLOPs in Fig.~\ref{fig5}. We can see that our SMSR achieves the best PSNR performance with low computational cost.

	\noindent{\textbf{Qualitative Results.}}
	Figure~\ref{fig6} compares the qualitative results achieved on Urban100. Compared to other methods, our SMSR produces better visual results with fewer artifacts, such as the lattices in $img\_004$ and the stripes on the building in $img\_033$. We further tested our SMSR on a real-world image to demonstrate its effectiveness. As shown in Fig.~\ref{fig8}, our SMSR achieves better perceptual quality while other methods suffer notable artifacts.
	
	\section{Conclusion}
	\textcolor{black}{In this paper, we explore the sparsity in  image SR to improve inference efficiency of SR networks. Specifically, we develop a sparse mask SR network to prune redundant computation.}
	Our spatial and channel masks work jointly to localize redundant computation at a fine-grained level such that our network can effectively reduce computational cost while maintaining comparable performance. Extensive experiments demonstrate that our network achieves the state-of-the-art performance with significant FLOPs reduction and a speedup on mobile devices.
	
	\section*{Acknowledge}
	The authors would like to thank anonymous reviewers for their insightful suggestions. Xiaoyu Dong is supported by RIKEN Junior Research Associate Program. Part of this work was done when she was a master student at HEU.

	{\small
		\bibliographystyle{ieee_fullname}
		\bibliographystyle{unsrt}
		\bibliography{super-resolution,other-CV-fields,neural-network}
	}
	
\end{document}